\documentclass[10pt,twocolumn,letterpaper]{article}

\usepackage{cvpr}              
\definecolor{cvprblue}{rgb}{0.21,0.49,0.74}
\usepackage[pagebackref,breaklinks,colorlinks,allcolors=cvprblue]{hyperref}

\usepackage[accsupp]{axessibility}  
\usepackage{orcidlink}
\usepackage{balance}

\def\paperID{*****} 
\def\confName{CVPR}
\def\confYear{2026}

\title{Robust Deepfake Detection: Mitigating Spatial Attention Drift \\ via Calibrated Complementary Ensembles}

\author{
Minh-Khoa Le-Phan\orcidlink{0009-0005-9707-4026}\thanks{Both authors contributed equally to this research as co-first authors} \quad
Minh-Hoang Le\orcidlink{0009-0005-1501-8080}\footnotemark[1] \quad
Trong-Le Do\orcidlink{0000-0002-2906-0360}\thanks{Corresponding author} \quad
Minh-Triet Tran\orcidlink{0000-0003-3046-3041} \\
University of Science, VNU-HCM, Ho Chi Minh City, Vietnam\\
Vietnam National University, Ho Chi Minh City, Vietnam\\
{\tt\small \{lpmkhoa22, lmhoang22\}@apcs.fitus.edu.vn}\\
{\tt\small dtle@selab.hcmus.edu.vn, tmtriet@fit.hcmus.edu.vn}
}

\begin{document}
\maketitle

\begin{abstract}
Current deepfake detection models achieve state-of-the-art performance on pristine academic datasets but suffer severe spatial attention drift under real-world compound degradations, such as blurring and severe lossy compression. To address this vulnerability, we propose a foundation-driven forensic framework that integrates an extreme compound degradation engine with a structurally constrained, multi-stream architecture. During training, our degradation pipeline systematically destroys high-frequency artifacts, optimizing the DINOv2-Giant backbone to extract invariant geometric and semantic priors. We then process images through three specialized pathways: a \textbf{Global Texture stream}, a \textbf{Localized Facial stream}, and a \textbf{Hybrid Semantic Fusion stream} incorporating CLIP. Through analyzing spatial attribution via Score-CAM and feature stability using Cosine Similarity, we quantitatively demonstrate that these streams extract non-redundant, complementary feature representations and stabilize attention entropy. By aggregating these predictions via a calibrated, discretized voting mechanism, our ensemble successfully suppresses background attention drift while acting as a robust geometric anchor. Our approach yields highly stable zero-shot generalization, achieving \textbf{Fourth Place in the NTIRE 2026 Robust Deepfake Detection Challenge at CVPR}. Code
is available at \href{https://github.com/khoalephanminh/ntire26-deepfake-challenge}{https://github.com/khoalephanminh/ntire26-deepfake-challenge}.
\end{abstract}

\vspace{-5mm}
\section{Introduction}
\label{sec:intro}
\vspace{-2mm}

Recent advancements in deepfake detection prioritize zero-shot generalization to combat the rapid evolution of generative models. To avoid overfitting to specific generator fingerprints, the field has largely pivoted toward large-scale foundation models~\cite{effort, udd, dfd-fcg, patchdiscontinuity}, whose rich representational spaces provide strong zero-shot priors. However, most adaptations are evaluated on pristine academic datasets that fail to reflect real-world social media scenarios, where media endures severe compound degradations like lossy compression and downsampling~\cite{pmm, laanet}. Under these noisy conditions, standard Vision Transformers (ViTs)~\cite{vit} suffer from spatial attention drift, losing structural focus and incorrectly anchoring to complex background artifacts rather than the actual facial forgery. Furthermore, fine-tuning these global models on pristine datasets inherently exhibits a strong texture bias, causing them to miss semantically impossible but texturally smooth structural errors, such as logical inconsistencies.

To address these vulnerabilities, we propose a comprehensive forensic framework that integrates massive data scaling with a structurally constrained, multi-stream architecture.
Firstly, we scale our training across a diverse pool of 14 datasets to establish robust representational power. However, training exclusively on pristine data leaves models unprepared for in-the-wild conditions and highly susceptible to generator-specific overfitting. To bridge this domain gap, we subject the training pool to an extreme compound degradation pipeline. This explicitly simulates real-world noises while systematically destroying high-frequency cues, thereby optimizing the DINOv2 backbone~\cite{dinov2} to abandon fragile texture shortcuts~\cite{geirhos2020shortcut, frank2020leveraging} and extract invariant facial geometry~\cite{sbi}. Secondly, to resolve the spatial attention drift common in global ViTs under heavy noise, we propose a multi-stream architecture adapted via LoRA~\cite{lora}. This includes three specialized pathways: a \textbf{Localized Facial Stream} acting as a strict geometric anchor, a \textbf{Global Texture Stream} to evaluate macro-context, and a \textbf{Hybrid Semantic Fusion Stream} (incorporating a frozen CLIP backbone~\cite{clip}) to detect logical inconsistencies~\cite{ojha2023towards}. Aggregating these complementary signals yields highly stable zero-shot generalization, securing \textbf{Fourth Place} in the NTIRE 2026 Robust Deepfake Detection Challenge at CVPR~\cite{ntire26deepfake}.

Our main contributions are summarized as follows:
\begin{itemize}
    \item We introduce a foundation-driven scaling framework, combining a diverse 14-dataset training pool with an 18-operation compound degradation pipeline to explicitly neutralize texture shortcut learning~\cite{geirhos2020shortcut} and extract robust facial geometry~\cite{sbi}.
    \item We propose a complementary multi-stream architecture anchored by DINOv2~\cite{dinov2} that mitigates the spatial attention drift and inherent texture bias of standard ViTs by simultaneously evaluating local geometry, global context, and semantic integrity via CLIP.
    \item We provide rigorous visual and quantitative analyses---utilizing Score-CAM~\cite{scorecam} spatial attribution and embedding Cosine Similarity---to empirically validate that our architecture prevents severe attention drift and extracts strongly complementary forensic features.
\end{itemize}

\vspace{-1mm}
\section{Related Work}
\label{sec:related_work}
\vspace{-1mm}

\paragraph{Generalizable Detection and Foundation Models.}
While early detectors struggled with cross-dataset generalization, recent advancements have largely solved this for pristine images via representation learning techniques like self-blended images (SBI)~\cite{sbi}, token-level shuffling~\cite{udd}, and localized artifact attention~\cite{laanet}. Concurrently, the paradigm has shifted towards adapting large pretrained foundation vision models for forensics. Effort~\cite{effort} demonstrated the efficacy of orthogonal modeling on CLIP features, catalyzing a wave of foundation model fine-tuning approaches. Recent works leverage Facial Feature Guided Adaptation~\cite{dfd-fcg}, patch-discontinuity mining~\cite{patchdiscontinuity}, and multi-modal interpretable frameworks~\cite{m2f2-det} to achieve state-of-the-art zero-shot detection capabilities.

\vspace{-4mm}
\paragraph{Robustness to Real-World Degradation.}
Despite high cross-dataset accuracy on high-quality datasets, modern foundation-model-based detectors remain highly vulnerable to in-the-wild transmission artifacts. Recent evaluations, such as those by Practical Manipulation Model~\cite{pmm}, reveal that simple perturbations like blur, resizing, or lossy compression catastrophically degrade the predictive accuracy of current models. While architectures like LAA-Net~\cite{laanet} have begun evaluating robustness against quality degradation, handling compound, non-linear noise distributions remains a critical open challenge. To bridge this gap, our work introduces a comprehensive compound degradation engine that systematically destroys fragile high-frequency cues during training, forcing the network to rely on persistent geometric and semantic anomalies.

\begin{figure*}[t]
\centering
\includegraphics[width=.9\textwidth]{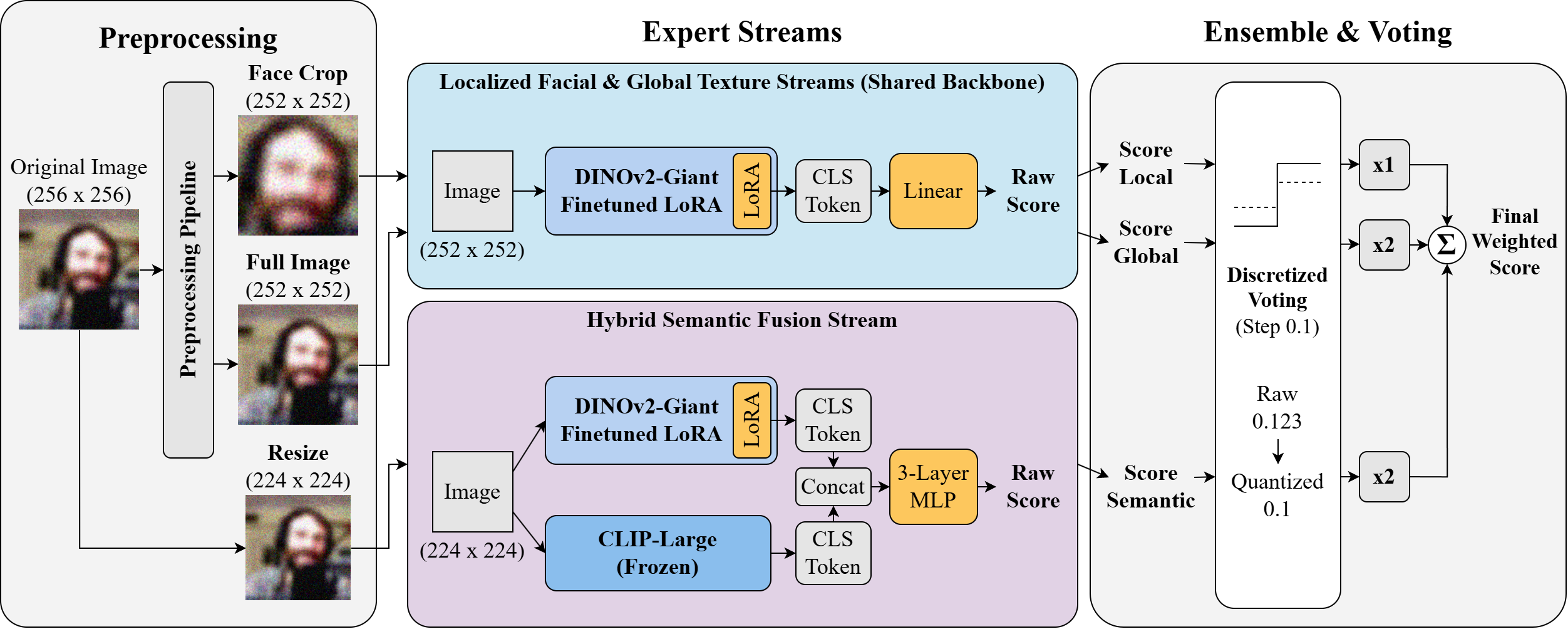}
\caption{\textbf{Overview of the proposed architecture.} The system processes inputs through three specialized expert streams. The \textit{Localized Facial} and \textit{Global Texture} streams maintain native signal integrity ($252\times252$) utilizing a shared DINOv2-Giant backbone~\cite{dinov2}. The \textit{Hybrid Semantic Fusion} stream ($224\times224$) concatenates geometric features from DINOv2~\cite{dinov2} with semantic features from a frozen CLIP-Large model~\cite{clip}. Trainable components (LoRA modules~\cite{lora} and MLPs) are highlighted in orange, while frozen/pretrained backbones are depicted in blue. Finally, raw probabilities are quantized to a 0.1 precision and aggregated via discretized probability voting (using a 1:2:2 weighting ratio for Local:Global:Fusion) to output a robust, calibrated final score.}
\label{fig:architecture}
\vspace{-4mm}
\end{figure*}

\vspace{-1mm}
\section{Proposed Method} 
\label{sec:method}
\vspace{-1mm}

The core intuition of our architecture is that real-world deepfake artifacts manifest across three distinct dimensions: local facial details, global image context, and high-level semantics. To capture these complementary features without overfitting to domain-specific noise, we propose three specialized expert streams trained under aggressive compound degradations (Figure~\ref{fig:architecture}). The \textbf{Localized Facial Stream} ($252 \times 252$) processes facial crops to isolate fine-grained manipulation traces, while the \textbf{Global Texture Stream} ($252 \times 252$) processes the full-frame context to capture broader spatial and contextual anomalies. Both visual streams use a DINOv2 foundation backbone~\cite{dinov2}, routing the final \texttt{[CLS]} token through a linear classification head. To catch complex logical errors that evade pure texture analysis, the \textbf{Hybrid Semantic Fusion Stream} ($224 \times 224$) acts as a safety net, leveraging the vision-language priors of CLIP~\cite{clip} to verify broader semantic consistency. All individual streams are optimized via standard Binary Cross-Entropy (BCE) loss. Finally, we aggregate the predictions using a discretized probability voting strategy with a 1:2:2 weighting ratio (Local:Global:Fusion).


\vspace{-2mm}
\subsection{Foundation Backbone and Domain Balancing} 
\label{subsec:dataset}
\vspace{-1mm}

Standard deepfake detection pipelines often rely on intricate multi-objective loss functions~\cite{zhao2021multi}, specialized artifact-hunting modules~\cite{luo2021generalizing}, or complex pseudo-forgery generation algorithms~\cite{sbi}. However, manually engineering forensic priors to counter rapidly evolving generative models is inherently unscalable. Instead of relying on complex, manually engineered modules, we adopt a foundation-driven scaling approach. We utilize DINOv2-Giant~\cite{dinov2} as our primary visual engine. Trained via self-supervised patch comparison, DINOv2 extracts exceptionally dense, localized features that are naturally sensitive to the structural discontinuities of face-swapping. Because fully fine-tuning massive foundation models induces catastrophic forgetting - overwriting their valuable zero-shot priors - we adapt the backbone using LoRA~\cite{lora}. This parameter-efficient strategy preserves the model's generalized knowledge while explicitly tuning its attention toward forensic boundaries.

To prevent the network from memorizing domain-specific algorithmic noise~\cite{geirhos2020shortcut}, we curate a massive training pool from 14 diverse face forgery datasets (detailed comprehensively in Section~\ref{subsec:experiment-setting}). Because our target domain is face-swapping and facial reenactment, we systematically filter out Entire Face Synthesis data to strictly isolate structural blending boundaries. Furthermore, naive aggregation of these datasets would cause the network to overfit to massive, lower-quality collections while ignoring subtle, high-fidelity threats. To ensure uniform representation across baseline, in-the-wild, and highly deceptive forgery distributions, we aggressively downsample overrepresented datasets to match the scale of the most challenging modern benchmarks. This rigorous curation yields a balanced master pool of $377,343$ frames across $190,680$ unique identities, containing $52.67\%$ real and $47.33\%$ fake instances.

\vspace{-1mm}
\subsection{Extreme Compound Degradation}
\label{subsec:extreme-degradation}
\vspace{-1mm}

\begin{figure}[t]
    \centering
    \includegraphics[width=.9\linewidth]{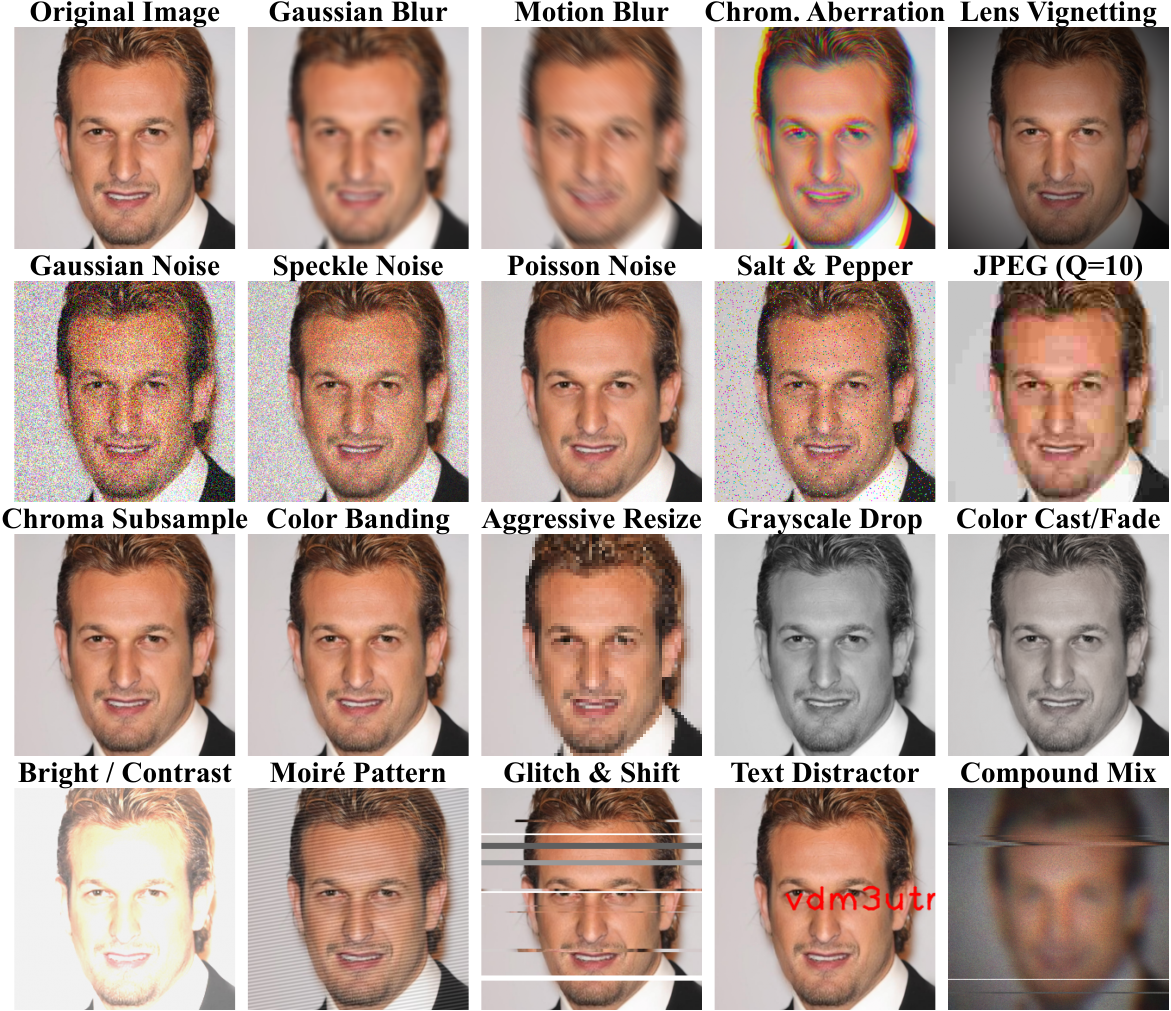}
    \vspace{-3mm}
    \caption{\textbf{Compound Degradation Engine.} Visual samples of the isolated degradation operations and an extreme compound mix (bottom right). This randomized pipeline simulates real-world transmission noise to explicitly neutralize texture shortcuts during training.}
    \label{fig:augmentation_grid}
    \vspace{-6mm}
\end{figure}

While massive data scaling provides necessary diversity, real-world deepfakes are rarely encountered in pristine, uncompressed formats. If trained exclusively on pristine images, Vision Transformers act as ``shortcut learners"~\cite{geirhos2020shortcut}, memorizing low-level generator noise rather than generalized structural reasoning. To explicitly prevent this, we draw inspiration from frameworks like the Practical Manipulation Model~\cite{pmm} and apply an extreme compound degradation pipeline to every training batch (Figure~\ref{fig:augmentation_grid}).

\vspace{-3mm}
\paragraph{The Degradation Engine.} Rather than applying fixed augmentations, our pipeline subjects each image to a randomized sequence of up to 15 degradation steps, drawn dynamically from a pool of 18 operations. The execution order is heavily shuffled, with independent application probabilities and dynamic severity strengths. To simulate the real-world cycle of users sharing and re-uploading media, we include a pre-processing loop that alternates between JPEG compression and resizing up to 5 times. The comprehensive operation pool covers four core categories: (1) \textbf{Compression and Resampling} (aggressive JPEG compression, chroma subsampling, color banding, and randomized multi-scale interpolation); (2) \textbf{Sensor and Digital Noise} (Gaussian, speckle, and Poisson noise, alongside simulated H.264 video glitches to mimic packet loss); (3) \textbf{Optical and Blur Artifacts} (anisotropic smoothing, motion blur, chromatic aberration, and vignetting); and (4) \textbf{Photometric Distortions and Distractors} (color casting, moiré patterns, and random text/patch overlays to enforce robustness under partial occlusion).

By subjecting the DINOv2 backbone to this severe real-world distribution, we effectively destroy easy texture shortcuts. This acts as a semantic forcing function, optimizing the model to extract the robust, geometric blending boundaries of the face.

\subsection{Mitigating Attention Drift and Texture Bias}
\label{subsec:attention-drift-and-texture-bias}

While foundation scaling and extreme degradation optimize the DINOv2 backbone for structural geometry, global Vision Transformers natively exhibit two critical vulnerabilities: spatial attention drift and semantic texture bias. To construct a truly robust ensemble, our architecture explicitly resolves these blind spots using specialized evaluation streams.

\vspace{-4mm}

\paragraph{Preventing Attention Drift (The Local and Global Streams).} 
Lacking spatial inductive biases~\cite{vit}, global ViTs frequently suffer from \textit{spatial attention drift}~\cite{han2025towards}, incorrectly anchoring to background noise rather than facial forgeries. We mitigate this by explicitly decoupling spatial evaluation into two parallel pathways. 

The \textbf{Localized Facial Stream} extracts a $1.3\times$ expanded facial crop to explicitly bound the model's attention. Because this strategy depends on accurate face localization, a major obstacle is that extreme compound degradation severely disrupts standard detectors (e.g., causing a 15\% RetinaFace~\cite{retinaface} failure rate on the challenge data). To ensure a stable geometric anchor, we deploy a 7-step preprocessing heuristic. Failed detections route through sequential recovery filters---bilateral filtering, heavy median blurring, GFPGAN~\cite{gfpgan} enhancement, sharpening CLAHE, non-local means, secondary median blurring, and NLMeans CLAHE (strict 0.9 confidence threshold)---to salvage viable geometry. This safeguard reduces the failure rate to just 1.8\%. Crucially, if facial extraction ultimately fails, this localized stream is bypassed.

Concurrently, because exclusively evaluating tight crops blinds the model to holistic environmental cues, the \textbf{Global Texture Stream} utilizes the shared DINOv2 weights to process the full-frame context. By separating these pathways, the architecture can safely capture macro-contextual anomalies---such as mismatched compression artifacts and spatial illumination inconsistencies---without allowing background noise to corrupt the localized facial attention.

To optimize these inputs for the DINOv2 backbone, which operates on $14 \times 14$ pixel patches, both streams process images at a $252 \times 252$ resolution. This dimension yields an exact $18 \times 18$ patch grid, preventing interpolation loss from standard $256 \times 256$ dimensions and strictly avoiding boundary padding artifacts. While the localized crop is resized to fit this grid, the global stream extracts a $252 \times 252$ center crop. This simple center-cropping strategy minimizes data loss and preserves native spatial frequencies better than full-image downsampling.

\vspace{-4mm}
\paragraph{Resolving Texture Bias (The Fusion Stream).} 
Despite robust data augmentation, DINOv2 remains inherently biased toward visual texture. This creates a vulnerability: if a generator synthesizes realistic skin but commits macro-semantic errors (e.g., blended earrings or distorted glasses), a purely texture-based evaluator may fail. To address this, we introduce the \textbf{Hybrid Semantic Fusion Stream} ($224 \times 224$). This pathway integrates a CLIP-Large backbone~\cite{clip}. Unlike vision-only models, CLIP is optimized via broad language supervision, enabling it to evaluate global logical consistency rather than just spatial frequency distributions~\cite{ojha2023towards}.

To preserve these semantic priors, the CLIP backbone remains entirely frozen. We extract the global \texttt{[CLS]} tokens from both the LoRA-adapted DINOv2 backbone and the frozen CLIP backbone, concatenate them, and project them through a 3-layer MLP. By keeping the CLIP weights fixed, we prevent modality co-adaptation~\cite{hinton2012improving} and mitigate the modality dominance common in joint training~\cite{wang2020makes}. This forces the trainable DINOv2 branch to learn complementary structural features rather than redundant semantic cues.

\vspace{-3mm}

\paragraph{Unified Ensemble Architecture.}
Ultimately, these three pathways form a mutually reinforcing ensemble. The Local stream anchors biometric geometry, the Global stream sweeps for macro-contextual anomalies, and the Fusion stream provides a safety net against semantic impossibilities. When evaluating unseen domains, raw continuous probabilities (e.g., exact float outputs like 0.814 and 0.842) often fluctuate slightly due to domain-specific noise rather than actual forgery features. To prevent this, we use a discretized voting mechanism. Instead of averaging the raw continuous scores, we first quantize the predictions to 0.1 precision steps (e.g., rounding both 0.814 and 0.842 into a single 0.8 confidence bin). This simple binning step discards tiny, meaningless variations, forcing the ensemble to aggregate based on broader, more stable confidence levels. As empirically validated in Section~\ref{sec:experiment} (Table~\ref{tab:ensemble_ablation}), this early-discretization strategy strictly outperforms continuous averaging. By calibrating these complementary perspectives, the final architecture mitigates the blind spots of individual foundation models, establishing a robust defense against compound degradations in the real world.

\vspace{-1mm}
\section{Experiments} \label{sec:experiment}
\subsection{Experimental Settings} \label{subsec:experiment-setting}
\vspace{-1mm}

\paragraph{Datasets and Balancing.} 
Our training pool integrates 14 curated datasets across four functional groups: (1) \textit{Baseline Single-Domain} (FaceForensics++~\cite{ff++}, UADFV~\cite{uadfv}); (2) \textit{Cross-Generator Diversity} (Celeb-DF-v2~\cite{cdfv2}, Celeb-DF-v3~\cite{cdfv3}, DeepFakeDetection~\cite{dfd}, DFDC~\cite{dfdc}, DFDCP~\cite{dfdcp}, FaceShifter~\cite{faceshifter}, DeeperForensics-1.0~\cite{deeperforensics10}); (3) \textit{In-the-Wild Context} (DDL~\cite{ddl}, DF40~\cite{df40}, FFIW~\cite{ffiw}); and (4) \textit{Modern High-Quality} (HIDF~\cite{hidf}, RedFace~\cite{redface}). To prevent overfitting to overrepresented classes, we employ a dynamic sampling strategy (8 frames per video) and stochastic item dropping. We explicitly filter out Entire Face Synthesis media from DF40 and RedFace to isolate face-swapping and reenactment artifacts. This yields a balanced master pool of 377,343 frames across 190,680 unique identities (52.67\% real, 47.33\% synthetic). To guarantee strict zero-shot evaluation, the official NTIRE challenge datasets were entirely isolated from our training phase. All images in these challenge sets are natively provided at a standard $256 \times 256$ resolution. During development, only the NTIRE Train (1,000 images; 500 real, 500 fake) contained accessible labels, and it was utilized strictly for validation. Consequently, our final performance is measured against the unseen NTIRE Validation (100 images; 50 real, 50 fake), NTIRE Public Test (1,000 images; 500 real, 500 fake), and NTIRE Private Test (unknown split), ensuring that our reported metrics reflect true domain generalization rather than localized overfitting.

\vspace{-3mm}
\paragraph{Implementation Details.}
All models were trained on a single NVIDIA A100 (80GB) GPU using AdamW optimization with a learning rate of $1\times 10^{-4}$ over approximately 10,000 iterations with a batch size of 32. To efficiently adapt the DINOv2-Giant backbone without overwriting pre-trained priors, we applied LoRA ($r=32$, $\alpha=64$, dropout $0.15$) targeting the \texttt{query}, \texttt{key}, \texttt{value}, and \texttt{dense} projection layers. The Localized and Global streams processed the facial crops and global center crops, respectively, at $252 \times 252$ resolution, while the Hybrid Semantic Fusion stream operated at CLIP ViT-L/14's native $224 \times 224$. The entire network was optimized using standard Binary Cross-Entropy (BCE) loss.

\vspace{-3mm}
\paragraph{Evaluation Measures.} 
Following the official NTIRE challenge protocols, model robustness and zero-shot generalization are evaluated using the Area Under the Receiver Operating Characteristic Curve (ROC-AUC). Additionally, the challenge guidelines strictly required all submitted probabilities to be quantized to a 0.1 precision.

\subsection{Performance on Hidden Domain Benchmarks}
\vspace{-1mm}

\begin{table}[b!]
\centering
\begin{tabular}{|c|c|c|c|}
\hline
Index & Model & public test & private test \\
\hline
(1)  & ShallowReal         & 0.9218          & \textbf{0.9168} \\
(2)  & INTSIG              & 0.8901          & 0.8824          \\
(3)  & AntInternational    & \textbf{0.9234} & 0.8691          \\
(4)  & \textbf{HCMUS-Aqua (Ours)} & 0.8775          & 0.8523          \\
(5)  & acvlab              & 0.8461          & -               \\
(6)  & Reagvis Lab         & 0.8430          & -               \\
(7)  & Hit Virlab          & 0.8418          & -               \\
(8)  & \textit{Anonymous}  & 0.8390          & -               \\
(9)  & Zeke                & 0.8240          & -               \\
(10) & TCD Vision          & 0.8045          & -               \\
(11) & PSU                 & 0.7056          & -               \\
(12) & AI4Good             & 0.6865          & -               \\
(13) & Acube               & 0.6860          & -               \\
(14) & NTR                 & 0.5999          & -               \\
\hline
\end{tabular}
\vspace{-2mm}
\caption{Comparison of our method with competitors in the final leaderboard (adopted from~\cite{ntire26deepfake}). Note: Private test scores for Teams (5) through (14) are unavailable (-) as the challenge organizers strictly reproduced and evaluated only the top 4 submissions to establish the final verified rankings.}
\label{tab:model_performance_basic}
\vspace{-0mm}
\end{table}

Table~\ref{tab:model_performance_basic} benchmarks our 1:2:2 calibrated ensemble against all officially verified submissions in the NTIRE 2026 Deepfake Detection Challenge. Our multi-stream approach secured a top-tier global rank, achieving an AUC of 0.8775 on the Public test and 0.8523 on the strictly hidden Private test.
Furthermore, our architecture establishes a significant margin over the competition, outperforming the teams outside the top 4 by over $3\%$ AUC.



\subsection{Ablation Studies}
\label{subsec:ablations}

To rigorously validate our architectural and experimental design choices, we conduct extensive ablation studies. We systematically isolate the contributions of our dataset composition, foundation backbone adaptations, degradation engine, and ensemble voting mechanism.

\vspace{-3mm}
\paragraph{Dataset Scaling Ablation.}

\begin{table*}[htbp]
\centering
\setlength{\tabcolsep}{6pt}
\begin{tabular}{|c||c|c|c|c||cc|}
\hline
& \multicolumn{4}{c||}{\textbf{Training Pool Composition}} & \multicolumn{2}{c|}{\textbf{AUC ($\uparrow$)}} \\
\cline{2-7}
\textbf{Phase} & Single-Domain & Cross-Generator & In-the-Wild & High-Quality & \textbf{val} & \textbf{public test} \\
& (Baseline) & (Diversity) & (Context) & (Modern) & & \\
\hline\hline
1 & \checkmark & & & & 0.7452 & - \\
2 & \checkmark & \checkmark & & & 0.8314 & - \\
3 & \checkmark & \checkmark & \checkmark & & \underline{0.9101} & \underline{0.8540} \\
\textbf{4 (Ours)} & \checkmark & \checkmark & \checkmark & \checkmark & \textbf{0.9303} & \textbf{0.8713} \\
\hline
\end{tabular}
\vspace{-2mm}
\caption{\textbf{Dataset Scaling Ablation.} Performance comparison of independent models trained on progressively expanding dataset combinations. Systematically expanding the pool from single-domain boundaries (Config 1) to encompass diverse rendering pipelines, uncontrolled contexts, and modern deceptive forgeries (Config 4) explicitly forces the network to abandon shortcut learning, maximizing zero-shot generalization on the degraded test set.}
\label{tab:data_scaling}
\vspace{-4mm}
\end{table*}

Table~\ref{tab:data_scaling} evaluates independent models trained from foundation weights on progressively expanding dataset groups (defined in Section~\ref{subsec:experiment-setting}) to validate our scaling strategy under zero-shot conditions. A baseline trained on \textit{Single-Domain} datasets (Config 1) yields a 0.7452 validation AUC, as it memorizes specific generator noise. Expanding the pool to include \textit{Cross-Generator Diversity} (Config 2) explicitly prevents these algorithmic shortcuts, increasing validation AUC to 0.8314. Incorporating \textit{In-the-Wild Context} datasets (Config 3) pushes validation AUC to 0.9101---and 0.8540 on the Public test set---by optimizing the model to ignore complex backgrounds and varied illumination. Finally, integrating \textit{Modern High-Quality} forgeries (Config 4) forces the network to isolate subtle structural blending anomalies. Yielding the highest performance across both validation (0.9303 AUC) and the Public test set (0.8713 AUC), this demonstrates that systematically structured data diversity is essential for zero-shot generalization under severe domain shift.

\vspace{-3mm}
\paragraph{Foundation Backbone and Tuning Strategy.} 
Table~\ref{tab:backbone_ablation} evaluates our foundation backbone selection and parameter-efficient tuning. DINOv2-Large (0.8376 AUC) outperforms CLIP-Large (0.8130 AUC), confirming that self-supervised patch reconstruction captures localized structural artifacts more effectively than global language supervision. Furthermore, while scaling to DINOv2-Giant increases parameter capacity, fully fine-tuning the network degrades performance (0.8255 AUC). Even when employing the two-phase Linear Probing then Fine-Tuning (LP-FT) protocol~\cite{lp-ft} to prevent initial gradient shock, full fine-tuning still induces catastrophic forgetting of the model's pre-trained zero-shot priors, leading to domain overfitting. Conversely, Low-Rank Adaptation (LoRA)~\cite{lora} strictly preserves these generalized weights while efficiently adapting the attention layers for forensic detection. This strategy yields the highest test AUC (0.8713), demonstrating that parameter-efficient tuning is mandatory for cross-domain robustness.

\begin{table}[htbp]
\centering
\begin{tabular}{|l|c|c|}
\hline
\textbf{Backbone \& Adaptation} & \textbf{val} & \textbf{public test} \\
\hline
CLIP-Large + LoRA & 0.8335 & 0.8130 \\
DINOv2-Large + LoRA & \underline{0.8974} & \underline{0.8376} \\
DINOv2-Giant + Full Finetune & 0.8519 & 0.8255 \\
\textbf{DINOv2-Giant + LoRA (Ours)} & \textbf{0.9303} & \textbf{0.8713} \\
\hline
\end{tabular}
\vspace{-2mm}
\caption{\textbf{Foundation and Tuning Ablation.} Parameter-efficient tuning (LoRA) prevents catastrophic forgetting, allowing the DINOv2-Giant backbone to significantly outperform both full fine-tuning and language-supervised models (CLIP) on the hidden test set.}
\label{tab:backbone_ablation}
\vspace{-3mm}
\end{table}

\begin{figure}[b!]
\centering
\vspace{-4mm}
\includegraphics[width=\linewidth]{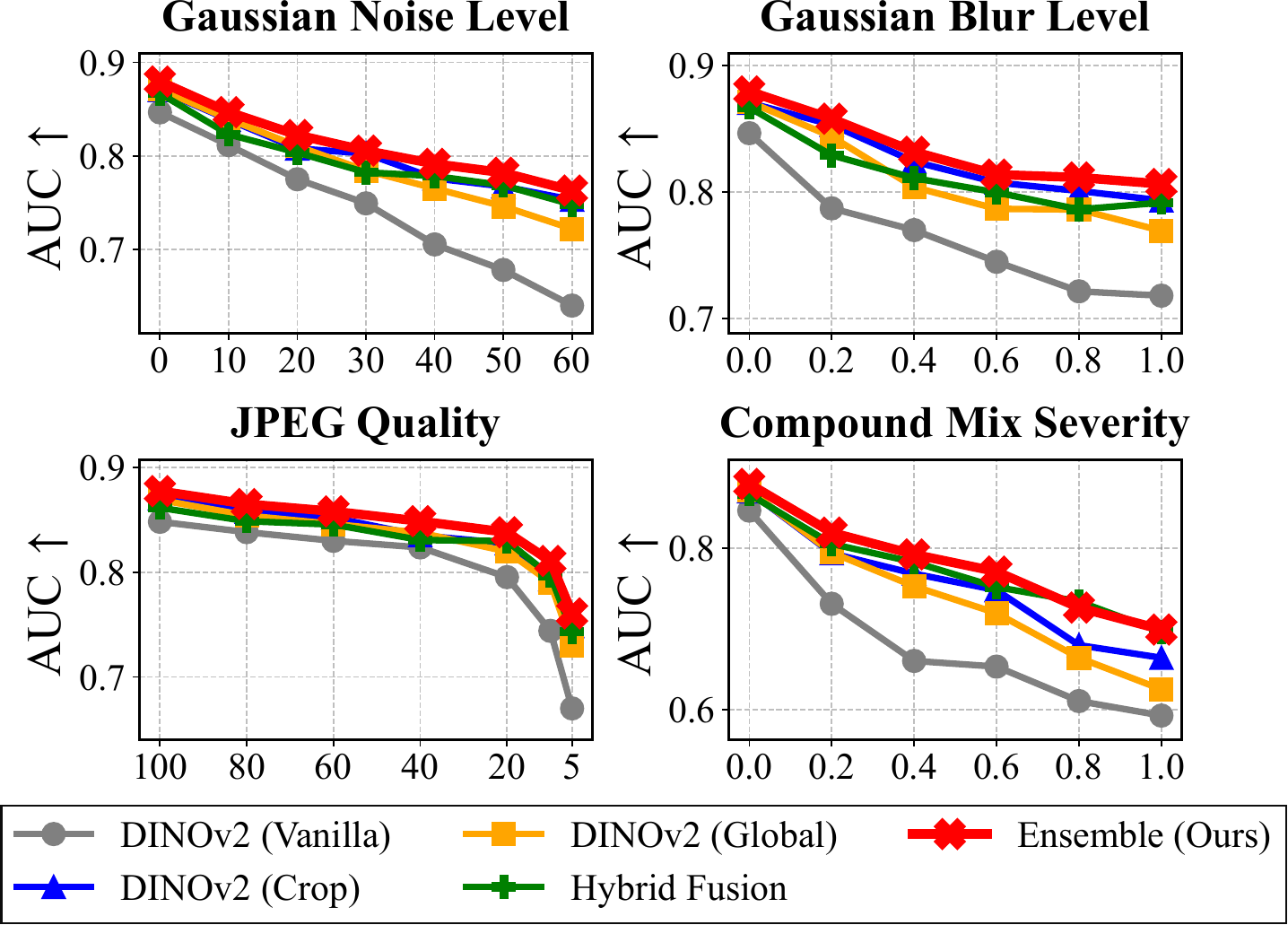}
\caption{\textbf{Complementary Ensemble Voting.} Evaluation of individual streams and aggregation strategies. All ensembles utilize discretized voting unless explicitly labeled as continuous. The calibrated 1:2:2 discretized ensemble yields the highest stability across both evaluation sets.}
\label{fig:auc_robustness}
\vspace{-3mm}
\end{figure}

\vspace{-4mm}
\paragraph{Compound Degradation and Scale Invariance.}
Table~\ref{tab:data_ablation} and Figure~\ref{fig:auc_robustness} validate our degradation and spatial alignment strategies. On the noisy Public test set, an unaugmented baseline - which processes the full 252x252 image through DINOv2 without our compound degradation engine (hereafter denoted as the Vanilla baseline) - collapses under isolated artifacts (Blur, Noise, JPEG) and drops below 0.60 AUC under compound noise. Conversely, our augmented ensemble maintains robustness, sustaining AUCs above 0.75 for maximum isolated severities and $\sim$0.70 under extreme compound degradation. Regarding resolution, downsampling standard $256 \times 256$ media to DINOv2's native $224 \times 224$ discards critical high-frequency cues (0.8589 AUC). By shifting the input to $252 \times 252$-yielding a perfect $18 \times 18$ grid of $14 \times 14$ patches-we minimize interpolation loss. The model adapts to this extended sequence length without positional embedding shock, successfully preserving structural integrity and unlocking our peak 0.8713 AUC.

\begin{table}[htbp]
\centering
\setlength{\tabcolsep}{4pt} 
\begin{tabular}{|l|c|c|}
\hline
\textbf{Configuration} & \textbf{val} & \textbf{public test} \\
\hline
Vanilla (252, No Aug) & 0.8422 & 0.8465 \\
Standard (224, Aug) & \underline{0.9248} & \underline{0.8589} \\
\textbf{Patch-Aligned (252, Aug)} & \textbf{0.9303} & \textbf{0.8713} \\
\hline
\end{tabular}
\vspace{-2mm}
\caption{\textbf{Degradation and Scale Invariance.} Extreme compound augmentation prevents shortcut learning, while strictly aligning the input resolution to the $252 \times 252$ patch grid minimizes interpolation loss.}
\vspace{-5mm}
\label{tab:data_ablation}
\end{table}

\begin{figure}[htbp]
    \centering
    \includegraphics[width=0.6\linewidth]{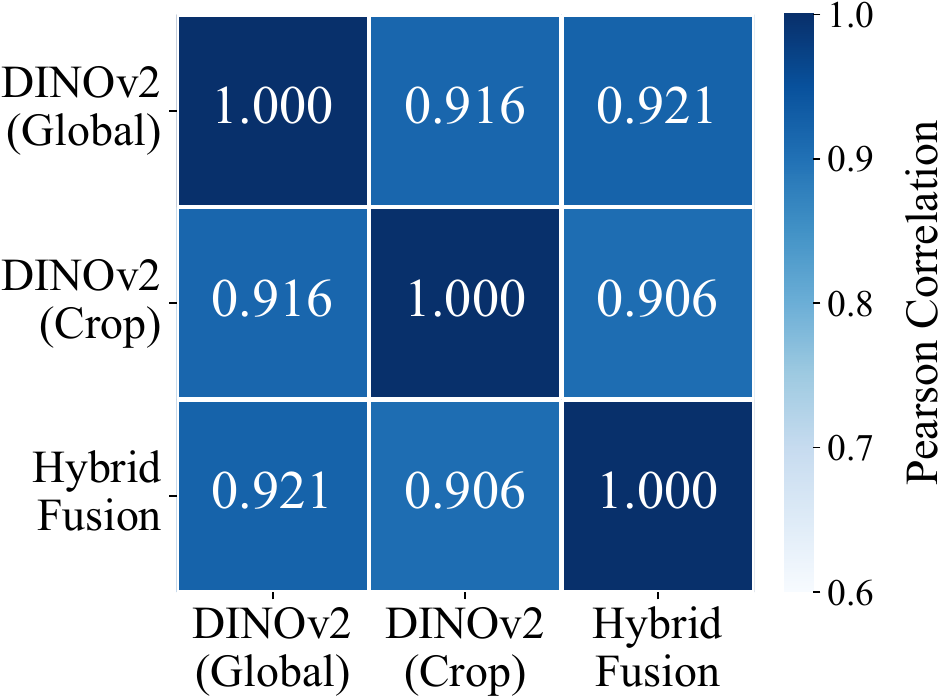}
    \vspace{-2mm}
    \caption{\textbf{Prediction Correlation Matrix.} While the streams naturally exhibit high correlation due to shared ground-truth targets, their strictly sub-unity values confirm that they do not redundantly collapse, but rather contribute complementary predictive signals to the ensemble.}
    \label{fig:correlation}
    \vspace{-4mm}
\end{figure}

\vspace{-4mm}
\paragraph{Complementary Ensemble Voting.}
Table~\ref{tab:ensemble_ablation} validates our multi-stream aggregation strategy. Ablating any single component strictly degrades performance, confirming all streams provide essential cues. To verify these streams learn distinct features, we analyze their pairwise prediction correlations (Figure~\ref{fig:correlation}). The sub-unity correlation values---Global-Crop (0.916), Global-Fusion (0.921), and Crop-Fusion (0.906)---demonstrate that the streams capture complementary decision boundaries, enabling the ensemble to smooth out domain-specific noise. Furthermore, Table~\ref{tab:ensemble_ablation} demonstrates the necessity of our 1:2:2 weighting calibration. Because extreme compound degradations frequently obscure high-frequency manipulation traces, our 1:2:2 ratio deliberately down-weights the Local stream in favor of the Global and Fusion streams, ensuring robustness even when localized facial features are heavily corrupted.

Finally, we compare our discretized voting against standard continuous averaging using identical weights (Table~\ref{tab:ensemble_ablation}, bottom rows). Quantizing predictions to 0.1 precision steps before aggregation improves both validation (0.9448 vs. 0.9220 AUC) and the public test set (0.8775 vs. 0.8773 AUC). This confirms that filtering out minor probability fluctuations creates a more robust consensus.

\begin{table}[t!]
\centering
\begin{tabular}{|l|c|c|}
\hline
\textbf{Ensemble Components} & \textbf{val} & \textbf{public test} \\
\hline
Localized Facial Stream & 0.8805 & 0.8705 \\
Global Texture Stream & \underline{0.9303} & 0.8713 \\
Hybrid Semantic Fusion Stream & 0.9170 & 0.8669 \\
\hline
Ensemble w/o local (0:1:1) & 0.9298 & 0.8745 \\
Ensemble w/o global (1:0:1) & 0.9072 & 0.8714 \\
Ensemble w/o semantic (1:1:0) & 0.9256 & 0.8762 \\
Unweighted Ensemble (1:1:1) & 0.9286 & 0.8764 \\
\textbf{Calibrated Ensemble (1:2:2)} & \textbf{0.9448} & \textbf{0.8775} \\
Continuous Ensemble (1:2:2) & 0.9220 & \underline{0.8773} \\
\hline
\end{tabular}
\vspace{-2mm}
\caption{\textbf{Complementary Ensemble Voting.} Evaluation of individual streams and aggregation strategies. All ensembles utilize discretized voting unless explicitly labeled as continuous. The calibrated 1:2:2 discretized ensemble yields the highest stability across both evaluation sets.}
\label{tab:ensemble_ablation}
\vspace{-3mm}
\end{table}

\begin{figure}[htbp]
    \centering
    \begin{minipage}{0.48\linewidth}
        \centering
        \includegraphics[width=\linewidth]{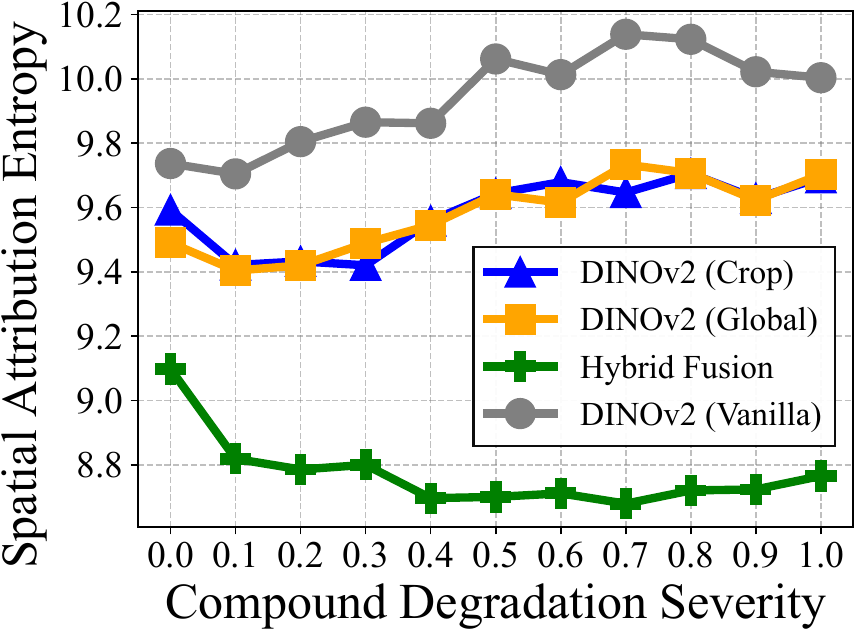}
    \end{minipage}\hfill
    \begin{minipage}{0.48\linewidth}
        \centering
        \includegraphics[width=\linewidth]{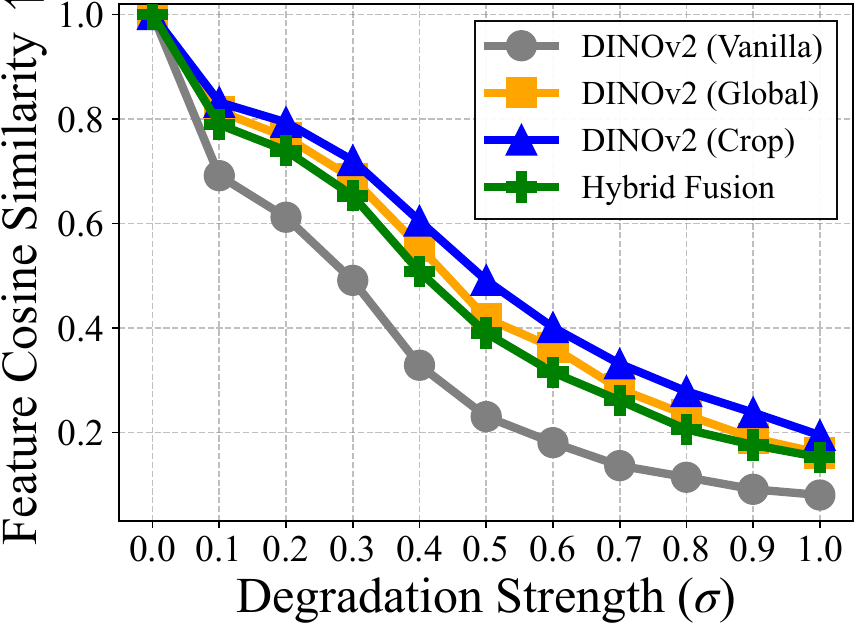}
    \end{minipage}
    \vspace{-2mm}
    \caption{\textit{Left:} Spatial Attribution Entropy, computed over normalized Score-CAM~\cite{scorecam} activation maps and averaged over all public test images. The unaugmented Vanilla baseline's attention scatters rapidly (high entropy). While the Crop and Global streams experience moderate drift, the Hybrid Fusion stream uniquely maintains a tightly localized, sub-9.0 focus. \textit{Right:} Feature Cosine Similarity, calculated between the final \texttt{[CLS]} token embeddings of the respective streams and averaged over all public test images. The Vanilla baseline's feature space collapses rapidly, whereas the augmented streams resist decay, with the localized Crop stream demonstrating the highest representational robustness.}
    \label{fig:xai_metrics}
    \vspace{-6mm}
\end{figure}

\vspace{-1mm}
\subsection{Visualizing Robustness and XAI}
\label{sec:xai}
\vspace{-1mm}

\begin{figure}[b!]
    \vspace{-3mm}
    \centering
    \includegraphics[width=0.9\linewidth]{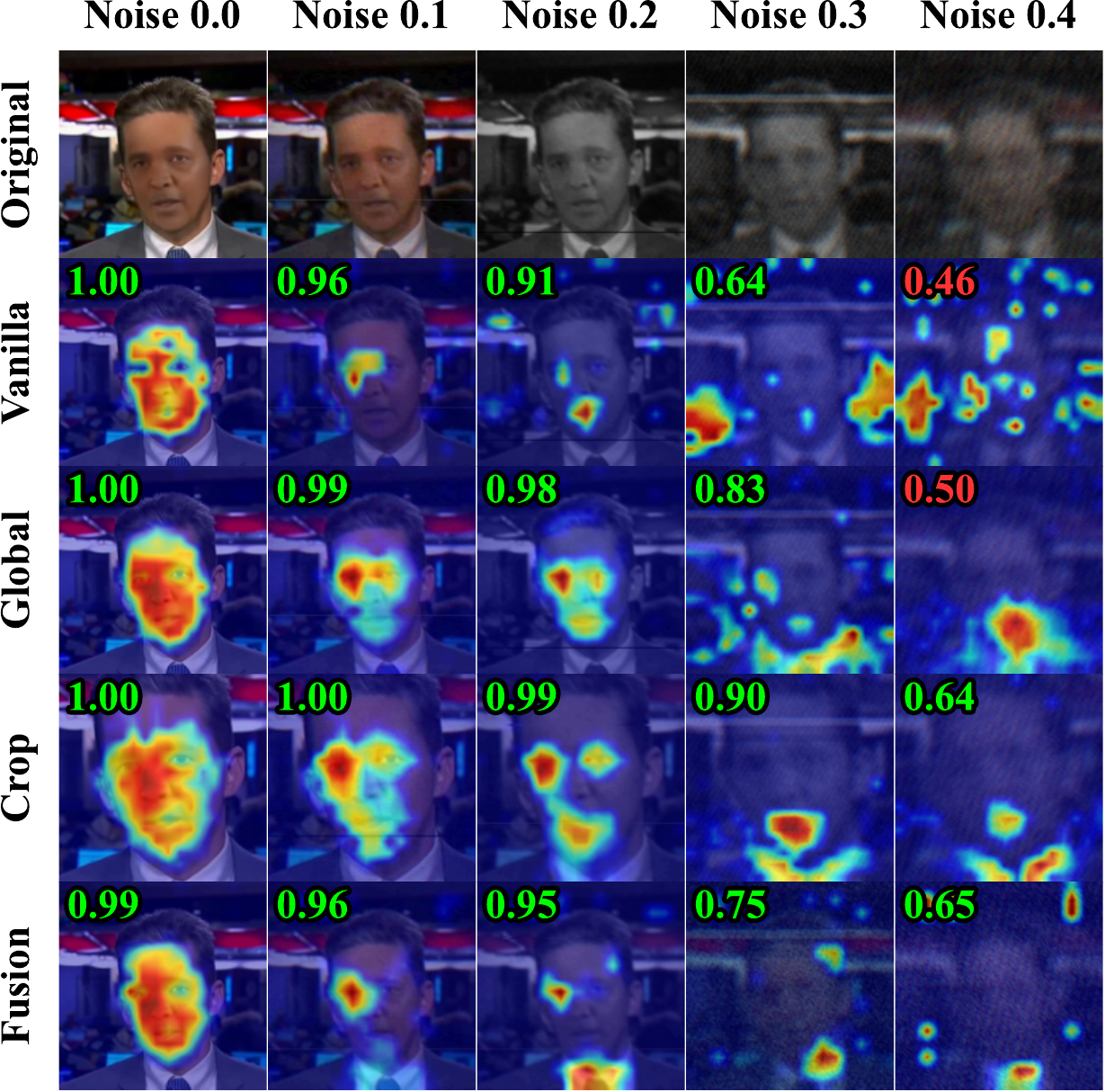}
    \vspace{-3mm}
    \caption{\textbf{Score-CAM~\cite{scorecam} Spatial Attribution under Compound Degradation.} Evaluated on a \textbf{Fake} Public test image (Top-left: forgery probability; Green=Correct, Red=Incorrect). At low noise (0.0--0.2, where 0.0 denotes the image's inherent noise), all models successfully localize the face. Under extreme noise (0.3--0.5), the Vanilla and Global models suffer severe attention drift, scattering attention into background static and yielding false negatives at 0.4. Conversely, our Crop and Fusion streams remain robust spatial anchors, preserving localized attention and correct classification.}
    \label{fig:hero_scorecam}
    \vspace{-5mm}
\end{figure}

\begin{figure*}[t]
    \centering
    \includegraphics[width=0.79\textwidth]{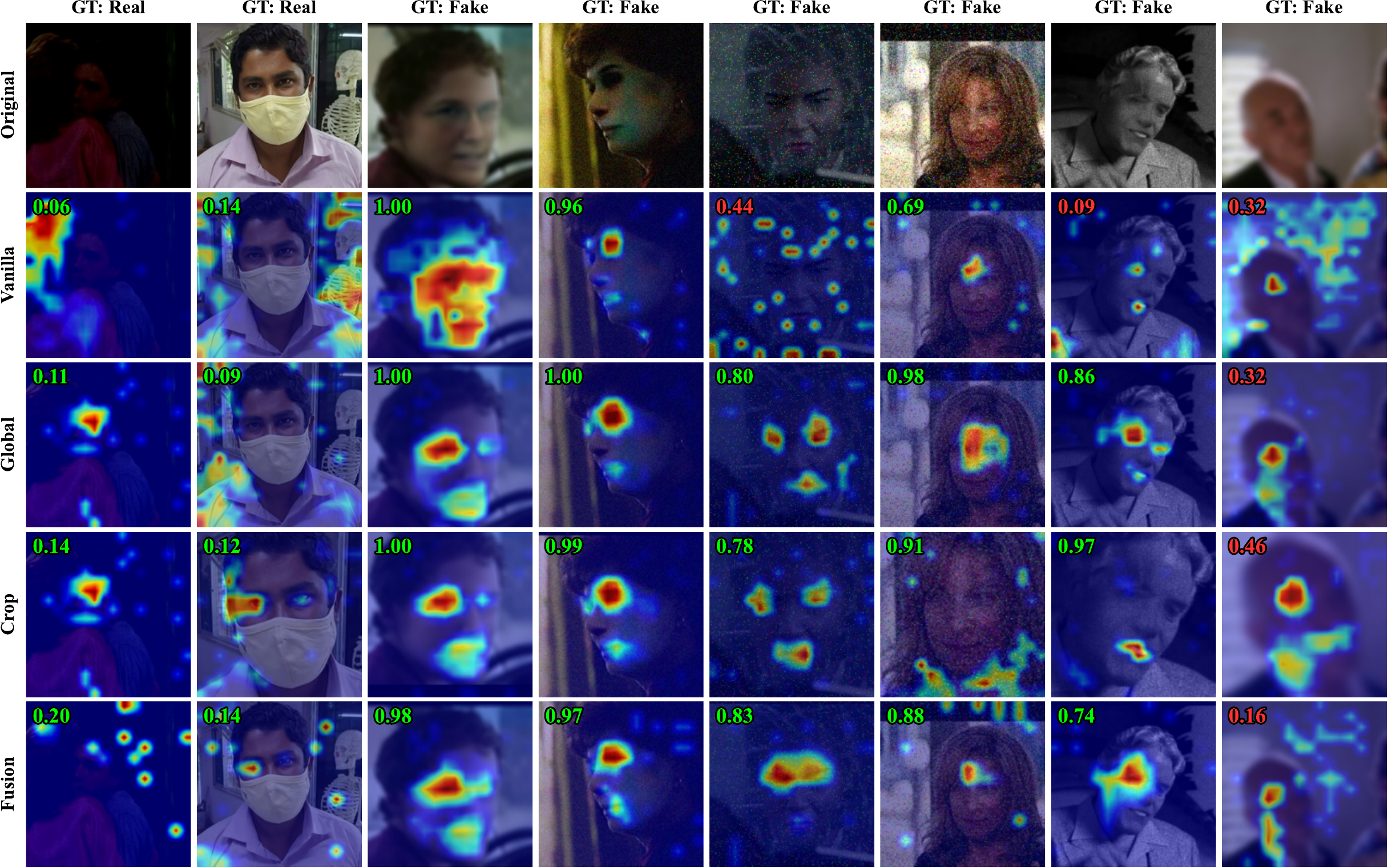}
    \vspace{-3mm}
    \caption{\textbf{Qualitative Score-CAM Analysis at Baseline Noise.} Evaluation across diverse challenge samples (Green = Correct prediction, Red = Incorrect). \textit{Left (True Negatives):} The ensemble successfully ignores complex, in-the-wild background distractors (e.g., clinical masks and skeletal models) without triggering false positive activations. \textit{Middle (True Positives):} On synthetic media, the multi-stream approach thrives; the Crop stream geometrically isolates spatial blending boundaries, while the Fusion stream acts as a semantic safety net, strictly attending to localized logical anomalies. \textit{Right (Limitations):} Extreme localized blur destroys high-frequency blending boundaries, forcing an honest failure across all models and highlighting a challenging boundary for current architectures.}
    \label{fig:scorecam_grids}
    \vspace{-5mm}
\end{figure*}

To physically validate the mechanisms behind our ensemble's zero-shot generalization, we map the internal feature representations using spatial attribution alongside feature stability metrics (Figures \ref{fig:xai_metrics}, \ref{fig:hero_scorecam}, and \ref{fig:scorecam_grids}). For spatial attribution, we explicitly utilize Score-CAM~\cite{scorecam}. Unlike standard gradient-based methods (e.g., Grad-CAM) which frequently suffer from gradient shattering~\cite{balduzzi2017shattered} and noisy saliency in deep Vision Transformers~\cite{chefer2021transformer}, Score-CAM relies purely on forward-pass activation scoring. This gradient-free approach yields highly precise, mathematically stable heatmaps that accurately reflect the true patch-level focus of the DINOv2 backbone without introducing backward-pass artifacts.

\vspace{-4mm}
\paragraph{Quantifying Attention Drift and Feature Stability.}
To strictly measure spatial attention drift under domain shift, we calculate Spatial Attribution Entropy. By treating the normalized Score-CAM~\cite{scorecam} activation map as a 2D probability distribution $P$, the entropy is computed as $-\sum P_{i,j} \log P_{i,j}$. High entropy indicates that the model's attention has scattered indiscriminately across the image (entropy collapse), whereas low entropy denotes a dense, localized geometric focus. Similarly, to measure representational robustness, we compute the Feature Cosine Similarity between the \texttt{[CLS]} token embedding of a heavily degraded image and its pristine baseline. A rapid decay indicates that domain noise has destroyed the model's internal feature space. Both metrics are averaged over all public test images. As shown in Figure~\ref{fig:xai_metrics}, the unaugmented Vanilla baseline exhibits escalating spatial entropy and rapid cosine decay (plummeting below 0.3 by severity 0.5). While the Crop and Global streams experience moderate spatial drift under extreme noise, the Hybrid Fusion stream uniquely acts as a rigid semantic anchor, maintaining a flat, sub-9.0 entropy profile. Concurrently, all three augmented streams strongly resist feature collapse, with the localized Crop stream preserving the highest cosine similarity deep into the degradation spectrum.

\vspace{-4mm}
\paragraph{Qualitative Score-CAM Analysis.}
Figure~\ref{fig:hero_scorecam} visualizes model attention across progressive noise levels. At low noise (0.0 to 0.2), all models successfully localize the face. However, as degradation reaches extreme severity (0.4), the Vanilla and Global models suffer from severe attention drift. Their focus shifts entirely to background noise, causing false negatives (probabilities dropping to 0.46 and 0.50). Conversely, the Crop and Fusion streams remain anchored to the face, maintaining correct classifications (0.64 and 0.65). Crucially, guided by the frozen CLIP semantic prior, the Fusion stream's heatmaps at severity 0.4 do not just form generic facial masks; instead, they precisely isolate the specific structural artifacts causing the semantic inconsistency.

Furthermore, the qualitative Score-CAM grid (Figure~\ref{fig:scorecam_grids}) demonstrates the ensemble's complementary roles. On authentic media (true negatives), the streams successfully ignore complex in-the-wild distractors (e.g., medical skeletons and harsh shadows) without triggering false positive activations. On synthetic media (true positives), the complementary behavior is visually evident: the Crop stream isolates blending boundaries that confuse the Global model, while the Fusion stream attends to logical anomalies. Finally, we visualize a failure case: an extreme localized blur that completely obscures high-frequency blending edges can still fool the ensemble, indicating an avenue for future robust feature extraction.

\balance
\vspace{-3mm}
\section{Conclusion}
\label{sec:conclusion}
\vspace{-2mm}

In this paper, we introduce a foundation-driven ensemble to address the vulnerability of deepfake detectors to real-world compound degradations. By integrating a robust degradation engine with three structurally constrained pathways (Global Texture, Localized Facial, and Hybrid Semantic Fusion), we mitigate the dependency on dataset-specific artifacts. Visual attribution confirms that these streams extract strongly complementary forensic priors. The multi-scale visual streams successfully resist the severe attention drift common in standard foundation models, while the semantic pathway effectively overcomes their inherent texture bias. Aggregating these diverse signals via a calibrated 1:2:2 discretized voting mechanism yields highly stable zero-shot generalization. Ultimately, our framework establishes a robust benchmark for in-the-wild deepfake forensics, securing Fourth Place in the NTIRE 2026 Robust Deepfake Detection Challenge at CVPR.

\section*{Acknowledgments}
This research is funded by Vietnam National University, Ho Chi Minh City (VNU-HCM) under grant number DS.C2025-18-13.

The authors would like to acknowledge Saigon AI Hub for its support in providing infrastructure and resources.


{
    \small
    \bibliographystyle{ieeenat_fullname}
    \bibliography{main}
}

\end{document}